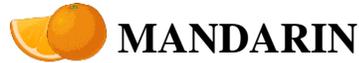

# MANDARIN

# Mixture-of-Experts Framework for Dynamic Delirium and Coma Prediction in ICU Patients: Development and Validation of an Acute Brain Dysfunction Prediction Model

Miguel Contreras [1,6], Jessica Sena[1,6], Andrea Davidson[3,6], Jiaqing Zhang[2,6], Tezcan Ozrazgat-Baslanti[3,6], Yuanfang Ren[3,6], Ziyuan Guan[3,6], Jeremy Balch[4,6], Tyler Loftus[4,6], Subhash Nerella[1,6], Azra Bihorac[3,6]*, Parisa Rashidi[1,6]**

[1]Department of Biomedical Engineering, University of Florida, Gainesville, FL, USA
[2]Department of Electrical and Computer Engineering, University of Florida, Gainesville, FL, USA
[3]Division of Nephrology, Department of Medicine, University of Florida, Gainesville, FL, USA
[4]Department of Surgery, University of Florida, Gainesville, FL, USA
[5]Department of Medicine, Stanford University, Stanford, CA, USA
[6]Intelligent Clinical Care Center (IC3), University of Florida, Gainesville, FL, USA

*Co-senior author

** Correspondence:

Parisa Rashidi
parisa.rashidi@ufl.edu

## Abstract

Acute brain dysfunction (ABD) is a form of organ dysfunction common in patients in the intensive care unit (ICU). This condition includes altered mental status, ranging from delirium to coma, and can lead to prolonged ICU stays, higher mortality, and long-term cognitive impairment. Traditional screening tools like Glasgow coma scale (GCS), confusion assessment method (CAM), and Richmond Agitation-Sedation Scale (RASS), aid diagnosis but rely on intermittent assessments, causing delays and inconsistencies. Although Artificial Intelligence (AI)-driven methods have been used for early delirium and coma prediction in the ICU, they often fail to capture the full spectrum of ABD, lack continuous monitoring, or do not leverage state-of-the-art AI techniques. In this study, we propose MANDARIN (Mixture-of-Experts Framework for Dynamic Delirium and Coma Prediction in ICU Patients), a 1.5M-parameter mixture-of-experts neural network to predict ABD in real-time among ICU patients. The model integrates ICU data from the time of admission (including vital signs, laboratory results, assessment scores, and medications) and patient characteristics (age, sex, race, and comorbidities) to predict the brain status in the next 12 to 72 hours. Our mixture-of-experts model can predict the next brain status based on the brain status at the time of prediction by leveraging a multi-branch approach. The MANDARIN model was trained on data from 92,734 patients (132,997 ICU admissions) from 2 hospitals between 2008–2019 and validated externally on data from 11,719 patients (14,519 ICU admissions) from 15 hospitals. Additionally, it was validated prospectively on data from 304 patients (503 ICU admissions) from one hospital in 2021–2024. Three datasets were used for training and evaluation: the University of Florida Health (UFH) dataset, the electronic ICU Collaborative Research Database (eICU), and the Medical Information Mart for Intensive Care (MIMIC)-IV dataset. MANDARIN significantly outperforms the baseline neurological assessment scores (GCS, CAM, and RASS) for delirium prediction in both external (AUROC 75.5% CI: 74.2%-76.8% compared to 68.3% CI: 66.9%-69.5%) and prospective (AUROC 82.0% CI: 74.8%-89.2% compared to 72.7% CI: 65.5%-81.0%) cohorts, as well as for coma prediction (external AUROC 87.3% CI: 85.9%-89.0% compared to 72.8% CI: 70.6%-74.9%, and prospective AUROC 93.4% CI: 88.5%-97.9% compared to 67.7% CI: 57.7%-76.8%) with a 12-hour lead time. This tool has the potential to assist clinicians in decision-making by continuously monitoring the brain status of patients in the ICU and predicting the risk of ABD in the following hours.

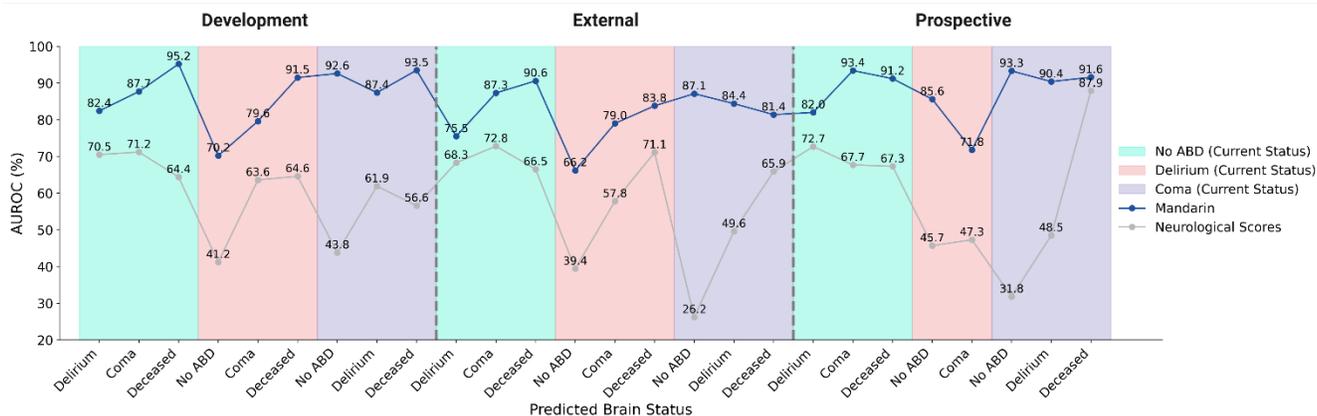

**Fig. 1 | MANDARIN performance for ABD prediction.** MANDARIN outperforms the clinical baseline (Neurological Scores) for the prediction of brain status in the ICU across three different cohorts.

## 1. Introduction

Acute brain dysfunction (ABD), encompassing delirium and coma, is a common and devastating complication in critically ill patients [1]. Delirium is the most frequent manifestation of ABD in the ICU, affecting up to 80% of mechanically ventilated patients [2]. Time spent in the ICU with delirium or coma is strongly associated with increased mortality and long-term cognitive deficits such as dementia [3]. For example, ICU delirium has been linked to a threefold higher risk of death, prolonged hospital stays, and greater likelihood of long-term cognitive impairment in survivors [4], [5]. These adverse outcomes underscore the critical importance of timely recognizing and managing ABD in the intensive care setting.

Currently, bedside neurological assessments are the mainstay for monitoring brain function in ICU patients. Tools like the Glasgow Coma Scale (GCS), Confusion Assessment Method for the ICU (CAM-ICU), and Richmond Agitation-Sedation Scale (RASS) are widely used to screen for coma and delirium states [6]. While these tools are validated and easy to administer, they have significant limitations. Assessments rely on subjective observations by clinical staff and are typically performed intermittently (e.g. once every few hours), providing only "snapshot" evaluations of a patient's neurologic status [7]. Important fluctuations in mental status can be missed between exams, leading to delays in delirium detection and inconsistent tracking of a patient's brain state. Indeed, it is estimated that less than half of delirium cases are actually recognized at the bedside with current standard methods [8]. In practice, ICU teams may not detect the onset of delirium or depth of coma until hours after it develops, which impedes early intervention.

Artificial intelligence (AI) offers a promising avenue to overcome these challenges through continuous, real-time brain monitoring. Modern Electronic Health Record (EHR) systems capture extensive data, including vital signs, medications, laboratory results, assessment scores, and demographic information, which can be leveraged with AI-driven methodologies to develop predictive models to identify ABD risk throughout a patient's ICU admission. Advanced deep learning techniques such as recurrent neural networks [9], transformer-based [10], and selective state space-based architectures [11] have demonstrated strong performance in forecasting various ICU-related outcomes, including acute kidney injury [12], acuity [13], and mortality [14]. These models excel in contextualizing high-dimensional data from EHR, a critical capability given the variability in ICU admissions, ranging from a few hours to several weeks. Recent innovations, have seen the use of mixture-of-experts (MoE) [15] in the domain of time series analysis (Time-MoE) [16]. These models have demonstrated robust performance in the use of time series data from different domains, while also demonstrating improved inference speed due to their sparsity [16]. These characteristics make it particularly suitable for processing the rich and complex EHR data associated with ICU monitoring, with the potential to predict ICU outcomes more accurately and effectively.

The application of AI models for predicting ABD in the ICU has been very limited. Most existing studies concentrate on predicting development of delirium. Several efforts have aimed at predicting the onset of delirium during an ICU admission [17], [18], [19], [20], postoperative delirium [21], or dynamically predicting delirium in ICU settings [22], [23]. To the best of our knowledge, only two studies have addressed prediction models that involve multiple ABD outcomes. The first study focused on forecasting next-day outcomes, including coma, delirium, ICU mortality, discharge, or normal brain function [24]. The second study applied a similar model to predict transitions from ABD states (including delirium and coma) to ABD-free conditions [25]. These studies did not utilize cutting-edge deep learning techniques such as transformers, selective state-space, or mixture-of-experts, Furthermore, these studies relied on small, private datasets collected from single cohort studies in individual hospitals, limiting reproducibility and robustness.

In this study, we propose MANDARIN (Mixture-of-Experts Framework for Dynamic Delirium and Coma Prediction in ICU Patients), a novel MoE model to predict real-time ABD in the ICU. MANDARIN uses EHR temporal data from the beginning of ICU admission along with static patient characteristic data to predict the risk of a patient developing ABD in the next 12 to 72 hours in the ICU. Specifically, the algorithm predicts the probability of a patient developing delirium or coma within the defined time windows, along with the probability of mortality. The MANDARIN model was developed using data from patients at the University of Florida Health (UFH) Shands Hospital between 2012–2019 in conjunction with data from the Medical Information Mart for Intensive Care (MIMIC)-IV [26]. The model was then externally validated on data from the eICU Collaborative Research Database (eICU) [27], and prospectively validated on data acquired prospectively from UFH [28] between May 2021 and August 2024. To our knowledge, the proposed MANDARIN model is the first real-time ABD prediction tool for patients in the ICU that incorporates the continuous prediction of delirium, coma, and mortality, while being extensively validated on three large datasets externally and prospectively.

## 2. Methods

2.1 Data and Study Design

This study utilized three datasets: UFH, MIMIC, and eICU, as illustrated in Fig. 2. The UFH dataset comprised both retrospective (UFH-R) and prospective (UFH-P) data, while the MIMIC and eICU datasets were exclusively retrospective. UFH-R includes data from adult ICU patients at the University of Florida Health Shands Hospital in Gainesville from 2012 to 2019. The UFH-P dataset encompasses records of adult ICU patients admitted between May 2021 and August 2024, gathered through a real-time data collection platform. The publicly accessible MIMIC dataset was sourced from the Beth Israel Deaconess Medical Center and covers 2008 to 2019 [26]. The eICU dataset represents ICU patient data from 208 U.S. hospitals in the Midwest, Northeast, South, and West regions during 2014–2015 [29].

Patients were excluded from all datasets if essential demographic data (age, discharge location, sex, race, or BMI), outcomes (*i.e.,* ABD states), or one of six standard vital signs—heart rate (HR), respiratory rate (RR), systolic blood pressure (SBP), diastolic blood pressure (DBP), body temperature (Temp), or oxygen saturation (SPO2)—were missing. Additionally, ICU stays shorter than 24 hours or exceeding 30 days were excluded. The lower threshold ensured sufficient data for at least one prediction and subsequent outcome validation, using a 12-hour observation window followed by a 12-hour prediction window. Furthermore, ICU admissions from hospitals in the eICU dataset that had less than 40% compliance for neurological scores (*i.e.,* GCS, RASS, and CAM missing in more than 60% of 12-hour shifts) were excluded. This criterion ensured that these scores were recorded roughly every 24 hours, allowing for more accurate ABD labels and minimizing potential false negatives (*i.e.,* missing ABD cases due to score missingness).

Data from UFH-R (2012–2019, 76,226 ICU admissions) and the MIMIC dataset (56,771 ICU admissions) formed the development dataset (132,997 ICU admissions), which was used for training and fine-tuning the MANDARIN model. External validation employed data from the eICU dataset (14,519 ICU admissions from 15 hospitals) to test the model's generalizability across different hospital settings. Prospective validation leveraged UFH-P data (2021-2024, 503 ICU admissions) to evaluate the model's application in a real-time clinical environment.

The research design encompassed six key stages: development, calibration, external validation, subgroup analysis, interpretability assessment, and prospective evaluation. This study adhered to the Transparent Reporting of a Multivariable Prediction Model for Individual Prognosis Or Diagnosis (TRIPOD) Statement [30], with the checklist available in the Supplementary Material. Initially, the MANDARIN model was trained and fine-tuned on the development dataset, dividing it into 90% for training and 10% for validation. Calibration involved sampling 10% of each validation dataset to align risk probabilities with the actual incidence rates of outcomes. The model underwent validation using data from multiple external hospitals to assess its generalizability. Subgroup analyses were conducted to assess performance differences based on age (18–60 versus over 60), sex, and race, alongside interpretability analyses to identify predictors most strongly associated with increased risk for ABD. The prospective evaluation was performed using real-time data collected through a platform interfaced with the Epic® EHR system [28], with retrospective retrieval for model assessment.

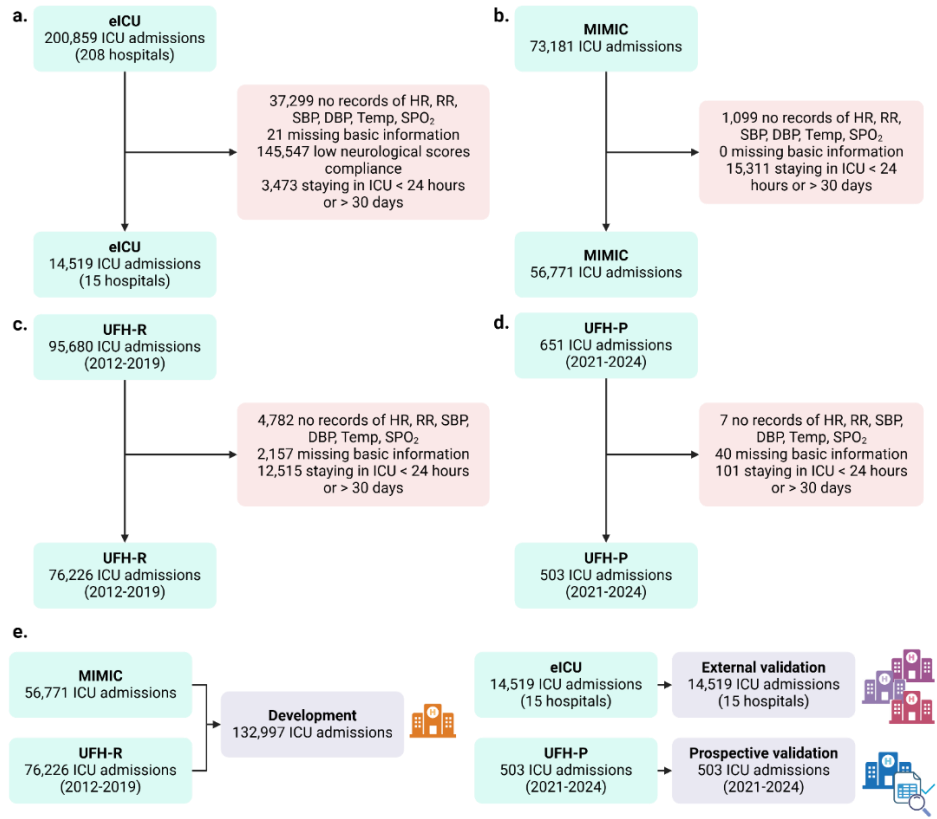

**Fig. 2 | Cohort flow diagram.** The **(a)** eICU, **(b)** MIMIC, **(c)** UFH-R, and **(d)** UFH-P datasets. **(e)** Final datasets were assembled from the four datasets for development (training and validation), external validation, and prospective validation. We created the development dataset by merging MIMIC and UFH-R datasets. Data from eICU dataset was used as external validation set. The prospective validation dataset was created from real-time data in the UFH-P dataset.

2.2 Ethics Approval and Patient Consent

Retrospective data from UFH (UFH-R) was obtained with the approval of the University of Florida Institutional Review Board and Privacy Office as an exempt study with a waiver of informed consent (IRB202101013). Prospective data from UFH was obtained by consent under the approval of the University of Florida Institutional Review Board under the numbers 201900354 and 202101013. Before enrolling patients in the study, written informed consent was obtained from all participants. In cases where patients could not provide informed consent, consent was obtained from a legally authorized representative (LAR) acting on their behalf. Eligible participants were individuals aged 18 and older who were admitted to the ICU and expected to remain there for at least 24 hours. Patients who could not provide an LAR or self-consent were expected to be transferred or discharged from the ICU within 24 hours, and those necessitating contact or isolation precautions were excluded. Also excluded from this study were patients who expired within 24 hours of recruitment. The eICU dataset is exempt from institutional review board approval due to the retrospective design, absence of direct patient intervention, and security schema.

The data in the MIMIC dataset is de-identified, and the institutional review boards of the Massachusetts Institute of Technology and Beth Israel Deaconess Medical Center both approved using the database for research.

2.3 Model Development

*2.3.1 Outcomes*

The primary outcome predicted by MANDARIN was the ABD state of a patient in the ICU in the next twelve hours obtained using computable phenotyping, defined by four levels of severity from the least to the most severe: no ABD, delirium, coma, and deceased [31]. Twelve-hour windows were chosen to capture the continuous transitions between ABD states that can happen in short time periods and to ensure lower score missingness [31]. Deceased state was determined through an indicator of whether a patient was discharged alive or passed away at the end of their stay in the ICU. The remaining three ABD states were determined through three neurological scores: GCS, CAM, and RASS. Specifically, for any 12-hour window: (1) coma was defined as at least one RASS score less than -3 or equal, or at least one RASS score equal to -3 along a GCS score less or equal to 8; (2) delirium was defined as at least one RASS score equal to -3 along a GCS greater than 8, or at least one RASS score greater than -3 along a positive CAM score; (3) no ABD was defined as not having met any of the previous conditions.

*2.3.2 Features*

In order to predict the outcome for the next twelve hours, static patient information and clinical events from ICU admission up to prediction time were used as predictive features. Clinical events consisted of the event's timestamp converted to hours since admission and time of the day, the variable code, and the value of the measured variable. In the case of multiple features happening at the same time point, they are arbitrarily ordered. Events are assigned to their corresponding observation window, where an observation window is comprised of all clinical events from the time of ICU admission until the start of the current prediction window. The prediction window represents a twelve-hour interval in which the outcome is assessed. In particular, vital signs, medications, laboratory tests, and assessment scores are extracted from ICU monitoring, while demographic and comorbidity information are extracted from patient history information. Variables occurring in less than 5% of all ICU stays are removed. Outliers are also removed based on impossible values based on clinical expertise and based on an upper (99th percentile) and lower (1st percentile) bound. A complete list of the variables used for prediction can be found in the Supplementary Material (Supplementary Table S1). Both temporal and static variables are then scaled according to their specific range using minimum-maximum scaling based on the minimum and maximum values for each feature in the development cohort. Finally, the MANDARIN model, which uses the embedding schema of a clinical time-series model [32] with incorporation of a mixture-of-experts model (MoE block, details in section 2.4) [16], is trained to predict probabilities for each ABD state and evaluated against ground truth labels. Given the inertia usually seen in clinical states (*i.e.*, the tendency to remain unchanged), there is an imbalance in the number of delirium and coma cases representing a transition. Therefore, to account for transitions between ABD states, 12-hour intervals were divided according to the 'current' status of the patient (*i.e.,* the ABD state at the time of prediction). The MANDARIN model was then branched according to the current status of the patient and predicted the probability of each ABD state different from the current ABD state. The complete workflow is summarized in Fig. 3.

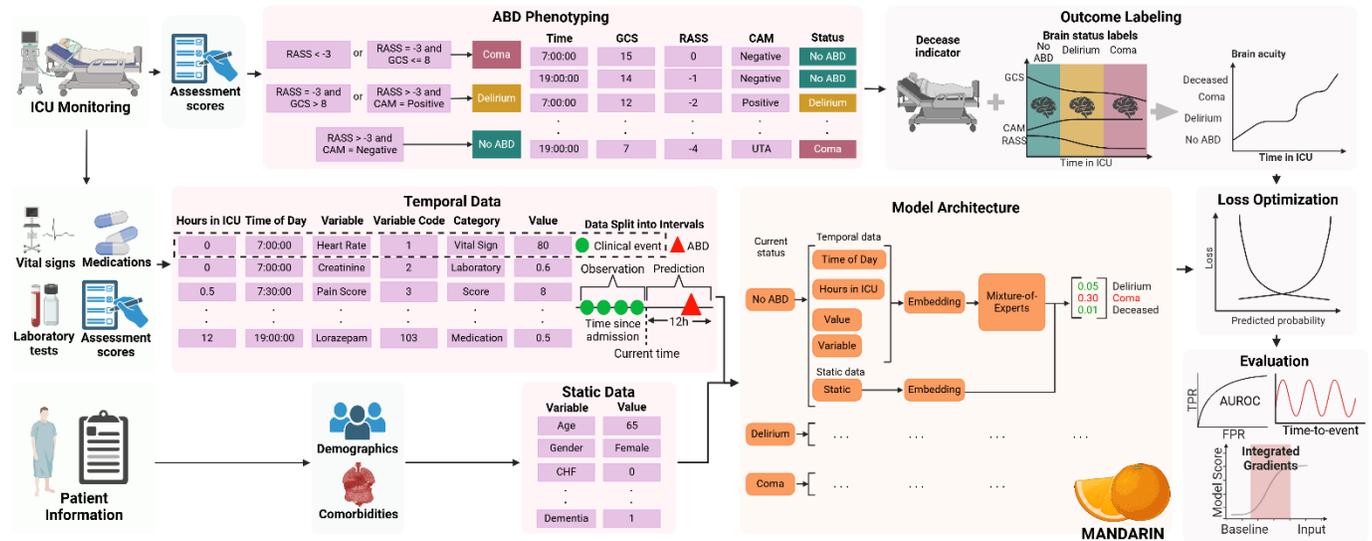

**Fig. 3: MANDARIN model development overview.** Data from four different Intensive Care Unit (ICU) datasets were used for training and validating the MANDARIN model (17 hospitals, 148,019 ICU admissions). Each dataset consisted of ICU monitoring data (*i.e.*, vital signs, medications, laboratory test results, and assessment scores) and patient admission information (*i.e.*, patient demographics and comorbidities). Three assessment scores were extracted for assessment of acute brain dysfunction (ABD) every 12 hours for every ICU admission: GCS, RASS and CAM. Coma was defined as a 12-hour interval in which the lowest RASS score was less than -3 or in which lowest RASS was equal to -3 and lowest GCS was less or equal to 8. Delirium was defined as a 12-hour interval in which the lowest RASS score was equal to -3 and lowest GCS was greater than 8 or in which lowest RASS score was greater than -3 and at least one positive CAM score. Temporal data was processed into a list of clinical events for each ICU admission, where each variable was assigned a code according to its order of appearance. Clinical events were assigned to observation windows spanning from the beginning of admission until the time of prediction and used with static data to predict the outcome in the next 12-hour window. The outcome for each 12-hour window was determined by combining ABD labels with decease indicators, providing a complete brain status trajectory for each patient. A mixture-of-experts model was employed for ABD prediction, with two separate embeddings used for temporal and static data. The first embedding was created by a Mixture-of-Experts block, which took the sequence of clinical events, from which the output was then combined with the second embedding, created from static patient data, to calculate the probabilities of each ABD outcome. The model was branched according to the current status (ABD state at the time of prediction) of the patient to account for transitions between states. The loss of the model was optimized by comparing the predicted probabilities with the ground truth labels. Finally, the model was evaluated by calculating the classification error for each outcome individually, performing time-to-event analysis to examine the difference in time from model alerts until an outcome of interest occurred and integrated gradients to understand features used by the model for prediction.

## 2.4 Model Architecture and Performance

### *2.4.1 Architecture*

After splitting the four datasets into development, external, and prospective, the development dataset was further split by randomly selecting 90% of the patients for training and 10% for validation and tuning hyperparameters. The MANDARIN model was developed based on the embedding schema of a clinical Transformer architecture [32], incorporating a mixture-of-experts block [16]. The original architecture was modified to create an embedding for temporal data using a sequence of time of recording (hours since admission), time of the day, value, and variable code. The time of recording, time of day, and value recordings were passed through one-dimensional (1D)-Convolutional layers, and the variable code recordings through an Embedding layer. The results were fused as one temporal embedding through addition, and positional encoding was incorporated to preserve the sequence order information. This approach removed the need for imputation that other time-series models require to maintain fixed dimensions in the input. The embedding schema presented here used padding/truncation to match the maximum sequence length set for the model, so no feature values were imputed. For static data, an embedding was created using two linear layers. A Mamba block was then employed to add context to the temporal embedding, from which the resulting vector was fused (i.e., added) with the static embedding. Finally, the fused vector was passed through individual multilayer-perceptron (MLP) modules, one for prediction of each ABD state. Each MLP outputs the probability of its corresponding ABD state across six time horizons: 12-hour, 24-hour, 36-hour, 48-hour, 60-hour, 72-hour. This approach allowed for earlier predictions of the outcomes (*i.e.,* more than twelve hours before) and accounted for cases where earlier signs of worsening ABD state were present. A cumulative distribution function (CDF) layer was applied for each MLP to maintain monotonicity of probabilities across time horizons, given that a higher probability of an outcome within a shorter time window implies a higher probability within a larger time window [12]. For training the model, class weights were applied to the loss function to account for the class imbalance for each ABD state and time horizon. The complete architecture is shown in Fig. 4.

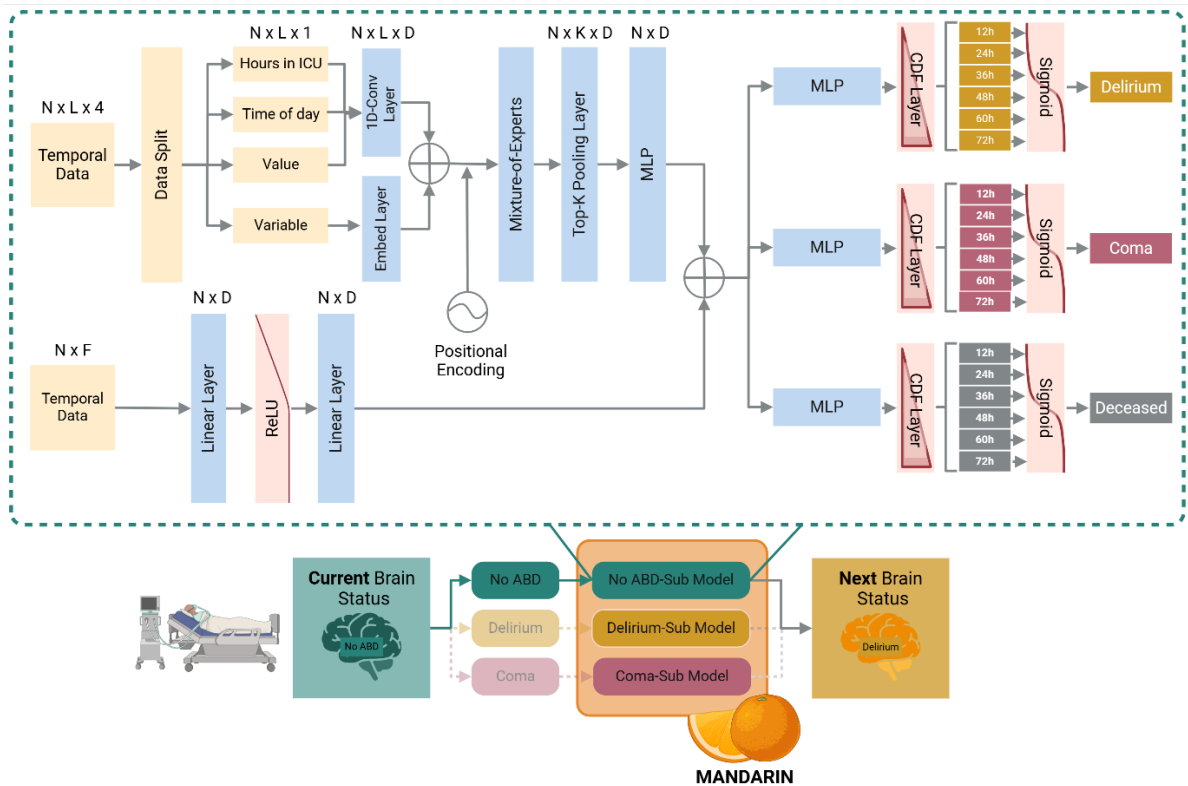

**Fig. 4 | MANDARIN model architecture.** Mixture-of-Experts model adapted to a clinical setting. The temporal data of dimensions $N$ (number of samples) $\times$ $L$ (sequence length) $\times$ 4 (quadruplet recording for each event) is split into individual matrices for hours since admission, time of day, value, and variable code of dimensions $N \times L$ each. The hours since admission, time of day, and value matrices go through a 1D-Convolutional Layer to create one embedding, while the variable code matrix goes through a look-up embedding layer to retrieve the embedding for the specific variable. Both embeddings were fused, and a positional encoding was added to conserve the sequence order information. The fused embedding was passed through a mixture-of-experts model, from which the last hidden state is taken and top-$k$ pooling is applied to retrieve the $k$-most important steps. The vector is then passed through a multi-layer perceptron (MLP) to create an $N \times D$ embedding fused with the static embedding resulting from passing the static data of dimensions $N \times f$ (static features) through two linear layers. Finally, the resulting vector is passed through three individual MLPs, which yield the probability for transition to each brain status outcome across six time horizons. A cumulative distribution function (CDF) layer is applied to keep monotonicity of probabilities across time horizons. The architecture is replicated on three branches depending on the current status (no ABD, delirium, coma) of the patient at the time of prediction.

*2.4.2 Baselines*

For baseline comparisons, three data processing methods were used: clinical, structured statistical features, and embedding. The clinical method involved using the three neurological scores used for ABD labelling (*i.e.,* GCS, RASS, and CAM) for prediction. Specifically, for each 12-hour interval, the maximum CAM score, and minimum RASS and GCS scores were taken and fused into one probability score (with a 0-1 range) for prediction. A lower RASS and GCS score, and a higher CAM score, meant an increase in the probability score. Although a RASS score higher than zero could mean a worsening state, for the purposes of this study, the lower RASS scores were considered as predictive of ABD, given they were indicative of delirium or coma according to the phenotyping logic used.

The structured statistical features method involved using statistical features from ICU monitoring data to represent the time-series from the ICU admission time to time of prediction in a static format, along with patient static information (*i.e.,* demographics and comorbidities). Four machine learning (ML) classification algorithms were tested for this method: logistic regression (LR) [33], random forest (RF) [34], extreme-gradient boosting (XGBoost) [35], and categorical boosting (CatBoost) [36].

The embedding method involved employing the embedding schema from MANDARIN (Section 2.4.1) with different deep learning (DL) methods. Four DL algorithms were tested for this method: long-short term memory (LSTM) [9], gated recurrent unit (GRU) [37], multi-head attention (Transformer) [10], and selective state space (Mamba) [38].

*2.4.3 Performance Evaluation*

For evaluation, six metrics were used: the Area Under the Receiver Operating Characteristic (AUROC), the Area Under the Precision-Recall Curve (AUPRC), sensitivity, specificity, Positive Predictive Value (PPV), and Negative Predictive Value (NPV). The Youden index [27] was used to find the optimal threshold to calculate sensitivity, specificity, PPV and NPV. Both AUROC and AUPRC were used as metrics to tune the model architecture and hyperparameters on the development test.

2.5 Calibration and Subgroup Analysis

Due to differences in the outcomes incidences (i.e., positive rates) in the four study cohorts, MANDARIN was calibrated using an isotonic regression algorithm [39] to improve the accuracy of the risk probability predictions. A random sampling of 10% for each validation set was used for model calibration. Calibration performance was evaluated by the Brier score and calibration curve and compared to all baseline models. To evaluate potential biases in prediction in different demographic groups, we conducted subgroup analysis by comparing model performance in age groups (young [18-65] and old [over 65]), gender groups (male and female), and racial groups (Black, White, and Others).

2.6 Delirium and Coma Phenotypes and Subtypes

The cause of delirium or coma episodes could be attributed to different factors, and on many occasions, individual episodes might be multi-factorial. Therefore, we divide delirium and coma into phenotypes and subtypes to evaluate predictions of these subgroups. Specifically, we divide delirium into five clinical phenotypes: hypoxic, septic, metabolic, sedative-associated, and unclassified. We based these phenotypes on previous work that has studied delirium phenotypes in critically ill patients and employ the same defining criteria [40]. We further divide delirium into two subtypes: hyperactive and hypoactive. We define hyperactive delirium as a positive CAM score along with a positive RASS score (*i.e.,* from 1 to 4), and hypoactive delirium as a positive CAM with a neutral or negative RASS score (*i.e.,* from 0 to -3) [41]. For coma, two subtypes were identified: induced and not induced. These subtypes were based on whether the coma was drug-induced or not, since these categories help distinguish between comas resulting from medical interventions, such as sedation for therapeutic purposes, and those caused by underlying medical conditions or injuries, which may have different prognostic and treatment implications.

2.7 Model Interpretability

For the interpretation of the MANDARIN model outputs, the integrated gradients method was used [42]. Integrated gradient attributions were calculated for each temporal variable by computing attributions from the embedding layer and the 1D-convolutional layer. To account for feature presence, we incorporate the proportion of ICU admissions in which a feature is present as part of the overall feature importance. Specifically, the following equation was used for calculating importance:

$$Importance(f) = p(f) \times (\alpha \times IGemb(f) + (1 - \alpha) \times IGconv(f)) \qquad (1)$$

Where $IGemb(f)$, $IGconv(f)$, and $p(f)$ are the attribution from the embedding layer, attribution for the 1D-convolutional layer, and proportion of feature $f$, and $\alpha$ is a hyperparameter that balances the contribution from the embedding versus the 1D-convolutional layer attributions. This method allowed computation of the importance of a feature across time steps, obtaining an overall score for each input feature to better understand risk predictors for each outcome.

2.8 Statistical Analysis

To determine if the difference in performance between baseline models and MANDARIN was statistically significant, all metric values between algorithms were compared using a Wilcoxon rank sum test. A 100-iteration bootstrap was performed to calculate the 95% confidence interval (CI) for each performance metric, and the median across the bootstrap was used to represent the overall value of each metric.

# 3. Results

3.1 Patient Characteristics

Patient characteristics for all cohorts in the study are shown in Table 1 in terms of ICU admissions. The eICU and MIMIC cohorts had the highest median age (65 years), while the UFH-R and UFH-P cohorts had slightly lower median ages (61 and 62 years, respectively). There were no significant differences in the percentage of female patients between the eICU and UFH-R cohorts (46.4% and 46.1%, respectively), but the UFH-P cohort had fewer female patients (38.8%) compared to all other cohorts. The UFH-P cohort had the highest ICU length of stay (6.5 days), while the MIMIC cohort had the shortest (2.4 days). The incidence of acute brain dysfunction varied across cohorts. Coma was most prevalent in the MIMIC cohort (39.0%) compared to eICU (17.9%), UFH-R (17.4%), and UFH-P (23.9%). Delirium was highest in the MIMIC cohort (26.2%) and lowest in UFH-R (15.6%). Mortality rates were highest in the MIMIC cohort (6.7%) and lowest in the UFH-P cohort (2.8%).

Table 1. Baseline characteristics for ICU admissions of four study cohorts.

| Item | | eICU (n = 14,519) | MIMIC (n = 56,771) | UFH-R (n = 76,226) | UFH-P (n = 503) |
|---|---|---|---|---|---|
| **Basic information** | | | | | |
| | Number of patients | 11,719 | 41,766 | 50,968 | 304 |
| | Number of hospital encounters | 13,479 | 52,256 | 72,032 | 444 |
| | Age, years, median (IQR) | 65.0 (54.0-76.0) [c,d] | 65.0 (54.0-76.0) [c,d] | 61.0 (49.0-71.0) [a,b] | 62.0 (50.0-71.0) [a,b] |
| | Female, n (%) | 6,735 (46.4%) [b,d] | 24,710 (43.5%) [a,c,d] | 35,111 (46.1%) [b,d] | 195 (38.8%) [a,b,c] |
| | BMI, kg/m$^2$, median (IQR) | 28.0 (23.8-33.5) | 27.5 (23.9-32.1) | 26.9 (23.0-32.1) | 26.7 (23.1-30.9) |
| | ICU length of stay, days, median (IQR) | 2.5 (1.7-4.1) [b,c,d] | 2.4 (1.6-4.3) [a,c,d] | 3.4 (2.0-6.2) [a,b,d] | 6.5 (3.3-11.7) [a,b,c] |
| | CCI, median (IQR) | 1.0 (1.0-2.0) | 3.0 (2.0-5.0) | 2.0 (0.0-4.0) [d] | 2.0 (1.0-4.0) [c] |
| **Race, n (%)** | | | | | |
| | Black | 2,804 (19.3%) [b,c] | 5,932 (10.4%) [a,c,d] | 15,381 (20.2%) [a,b,d] | 83 (16.5%) [b,c] |
| | White | 10,711 (73.8%) [b,c] | 38,667 (68.1%) [a,c,d] | 57,080 (74.9%) [a,b] | 386 (76.7%) [b] |
| | Other | 1,004 (6.9%) [b,c] | 12,172 (21.4%) [a,c,d] | 3,765 (4.9%) [a,b] | 34 (6.8%) [b] |
| **Comorbidities, n (%)** | | | | | |
| | CVD | 542 (3.7%) [c,d] | 2,288 (4.0%) [c,d] | 10,258 (13.5%) [a,b] | 53 (10.5%) [a,b] |
| | CHF | 749 (5.2%) [b,c,d] | 5,146 (9.1%) [a,c,d] | 20,339 (26.7%) [a,b] | 146 (29.0%) [a,b] |
| | COPD | 527 (3.6%) [b,c,d] | 4,830 (8.5%) [a,c,d] | 23,759 (31.2%) [a,b,d] | 124 (24.7%) [a,b,c] |
| | Dementia | 0 (0.0%) [b,c,d] | 82 (0.1%) [a,c,d] | 2,738 (3.6%) [a,b] | 17 (3.4%) [a,b] |
| | Renal disease | 444 (3.1%) [b,c,d] | 7,138 (12.6%) [a,c,d] | 16,516 (21.7%) [a,b,d] | 151 (30.0%) [a,b,c] |
| **Acute brain dysfunction, n (%)** | | | | | |
| | Delirium | 3,491 (24.0%) [b,c] | 14,850 (26.2%) [a,c] | 11,905 (15.6%) [a,b,d] | 113 (22.5%) [c] |
| | Coma | 2,602 (17.9%) [b,d] | 22,123 (39.0%) [a,c,d] | 13,228 (17.4%) [b,d] | 120 (23.9%) [a,b,c] |
| **Outcomes, n (%)** | | | | | |
| | Mortality | 606 (4.2%) [b] | 3,782 (6.7%) [a,c,d] | 3,000 (3.9%) [b] | 14 (2.8%) [b] |

Abbreviations: BMI: Body Mass Index; CCI: Charlson Comorbidity Index; CHF: Congestive Heart Failure; COPD: Chronic Obstructive Pulmonary Disease; CVD: Cerebrovascular Disease IQR: interquartile range.
P-values for continuous variables are based on a pairwise Wilcoxon rank sum test. P-values for categorical variables are based on pairwise Pearson's chi-squared test for proportions.
[a] p-value < 0.05 compared to the eICU cohort.
[b] p-value < 0.05 compared to MIMIC cohort.
[c] p-value < 0.05 compared to UFH-R cohort.
[d] p-value < 0.05 compared to UFH-P cohort.

## 3.2 State Transitions

The probability for each brain status transition across all study cohorts was calculated and is shown in Fig. 4. For patients with no ABD, there was a 91.81–97.24% probability of remaining without ABD, a 1.77–4.97% probability of transitioning to delirium, a 1.00–2.92% probability of transitioning to coma, and a 0.11–0.30% probability of transitioning to deceased. For patients with delirium, there was a 53.49–65.02% probability of remaining in delirium, a 28.36–39.84% probability of transitioning to no ABD, a 5.15–8.28% probability of transitioning to coma, and a 0.49–0.73% probability of transitioning to deceased. For patients in coma, there was a 64.52–66.87% probability of remaining in coma, a 19.33–23.79% probability of transitioning to no ABD, a 7.19–11.17% probability of transitioning to delirium, and a 0.81–3.29% probability of transitioning to deceased.

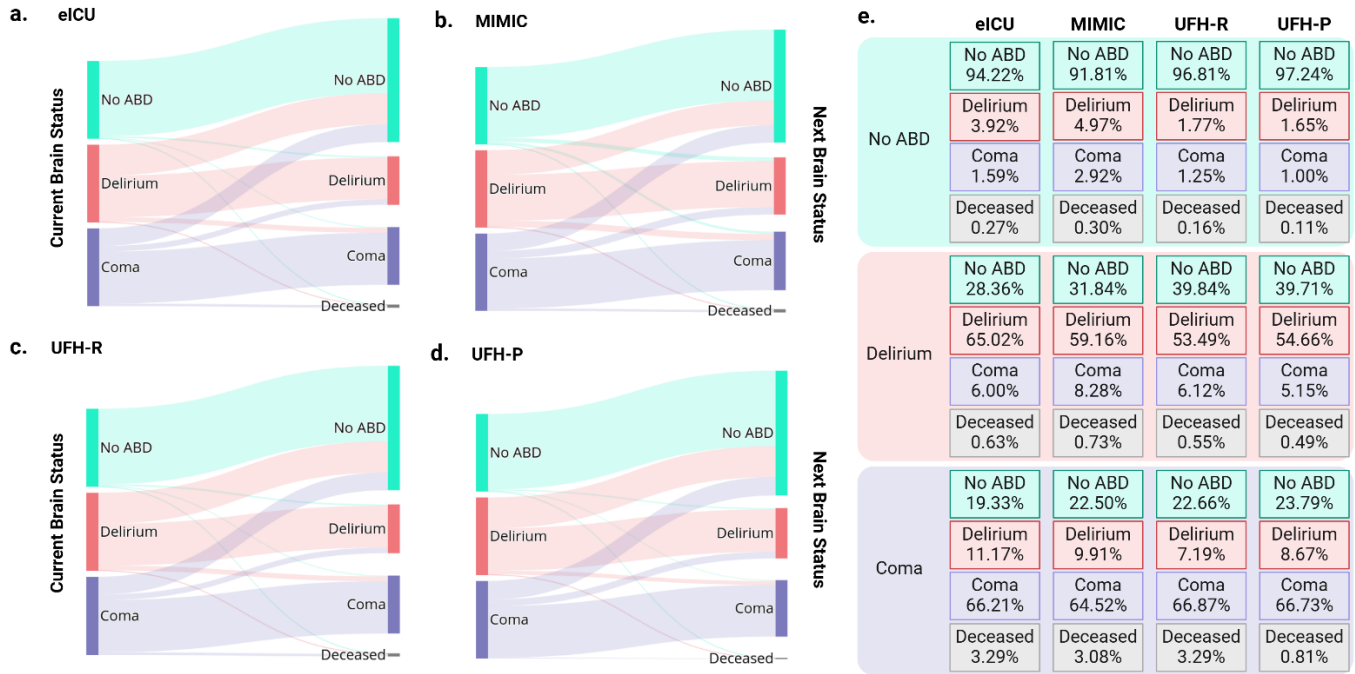

**Fig. 4: Brain status transition probability.** Sankey plots for brain status transition probability in: **(a)** eICU cohort, **(b)** MIMIC cohort, **(c)** UFH-R cohort, and **(d)** UFH-P cohort. **(e)** Table showing transition probability values for each transition on each cohort, The brain status transition probability is shown for every 12-h window. The probability was calculated based on the patient's current brain status (no ABD, delirium, or coma) and the probability to transition to each brain status. The probabilities for deceased were not calculated since it is an end outcome.

3.3 Model Performance, Calibration, and Subgroup Analysis

Bar plots in Fig. 5 illustrate the AUROC of the Mandarin model compared to baselines across different prediction tasks (delirium, coma, and deceased) and cohorts (development, external, and prospective) when current status is No ABD, with error bars representing 95% confidence intervals. Results are summarized by comparing Mandarin to the clinical baseline (Neurological Scores), the best ML model (CB), and the best DL model (Mamba). A comprehensive comparison of all models, as well as calibration and subgroup analysis, can be found in the Supplemental Material.

In the development cohort, Mandarin achieved AUROC values ranging from 82.4% (95% CI: 80.8%, 83.7%) to 79.7% (95% CI: 79.0%, 80.6%) for delirium, 87.7% (95% CI: 86.6%, 88.9%) to 81.5% (95% CI: 80.7%, 82.4%) for coma, and 95.2% (95% CI: 91.7%, 97.3%) to 91.8% (95% CI: 90.3%, 92.7%) for

deceased, as the prediction horizon increased from 12 to 72 hours. The second best model, Mamba, performed similarly, with AUROC values ranging from 82.7% (95% CI: 81.4%, 83.9%) to 80.3% (95% CI: 79.6%, 81.1%) for delirium, 87.7% (95% CI: 86.4%, 89.0%) to 82.1% (95% CI: 81.3%, 83.0%) for coma, and 95.3% (95% CI: 92.8%, 97.4%) to 91.5% (95% CI: 90.3%, 92.7%) for deceased.

In the external cohort, Mandarin achieved AUROC values ranging from 75.5% (95% CI: 74.2%, 76.8%) to 75.0% (95% CI: 74.2%, 75.6%) for delirium, 87.3% (95% CI: 85.9%, 89.0%) to 82.1% (95% CI: 80.9%, 83.2%) for coma, and 90.6% (95% CI: 87.8%, 93.0%) to 86.3% (95% CI: 84.8%, 88.0%) for deceased. Mamba showed similar performance, ranging from 75.4% (95% CI: 74.1%, 76.7%) to 75.0% (95% CI: 74.2%, 75.8%) for delirium, 86.8% (95% CI: 85.3%, 88.6%) to 81.2% (95% CI: 80.0%, 82.6%) for coma, and 87.4% (95% CI: 84.2%, 90.7%) to 82.1% (95% CI: 80.0%, 84.3%) for deceased.

In the prospective cohort, Mandarin's AUROC ranged from 82.0% (95% CI: 74.8%, 89.2%) to 71.4% (95% CI: 67.9%, 75.3%) for delirium, 93.4% (95% CI: 88.5%, 97.9%) to 75.4% (95% CI: 70.9%, 81.4%) for coma, and 91.2% (95% CI: 78.3%, 100.0%) to 93.8% (95% CI: 90.1%, 95.8%) for deceased. Mamba performed slightly lower, with AUROC values ranging from 81.1% (95% CI: 72.8%, 87.3%) to 72.0% (95% CI: 67.0%, 75.9%) for delirium, 93.8% (95% CI: 90.9%, 97.1%) to 74.4% (95% CI: 68.2%, 81.4%) for coma, and 75.5% (95% CI: 35.5%, 100.0%) to 79.4% (95% CI: 67.3%, 91.8%) for deceased.

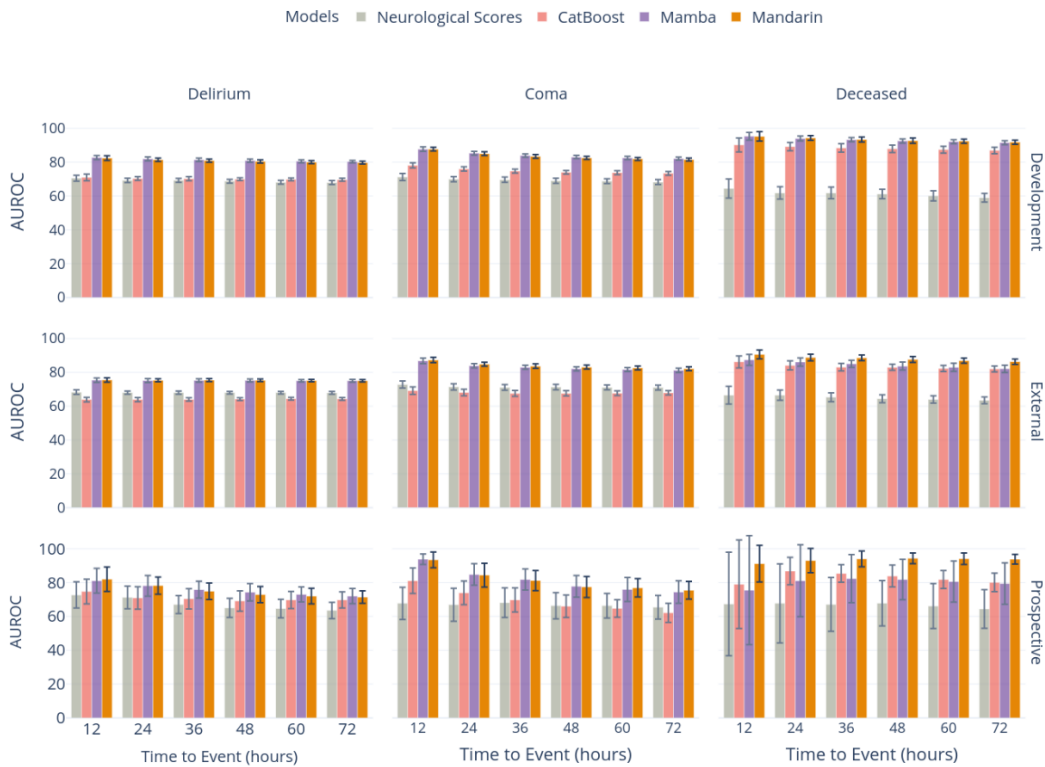

**Fig. 5: Model performance for predicting delirium, coma, and deceased across different cohorts and prediction windows when current status is No ABD.** Performance is measured using AUROC scores across the three validation cohorts: Development (top row), External (middle row), and Prospective (bottom row). The models compared include Neurological Scores (clinical baseline), CatBoost (best ML model), Mamba (best DL model), and Mandarin (proposed model). Error bars represent 95% CI.

3.4 Delirium and Coma Sub-Phenotypes

For a more comprehensive analysis of Mandarin's performance, we evaluate prediction performance for different clinical phenotypes of delirium, as well as delirium and coma subtypes (Fig. 6). The results showed that for delirium phenotypes, Mandarin and Mamba consistently outperformed the Neurological Scores and CatBoost in both external and prospective cohorts at the 12-hour prediction horizon. In the external cohort, Mandarin achieved AUROCs ranging from 82.4% (95% CI: 80.7%, 84.0%) for sedative delirium to 70.7% (95% CI: 67.4%, 73.7%) for unclassified delirium, while Mamba showed similar performance with AUROCs reaching 81.8% (95% CI: 80.0%, 83.3%) for sedative delirium to 69.1% (95% CI: 64.9%, 72.7%) for unclassified delirium. Neurological Scores and CatBoost achieved lower AUROCs, particularly in the unclassified category: 66.1% (95% CI: 63.1%, 69.1%) and 54.7% (95% CI: 51.8%, 58.3%) respectively. In the prospective cohort, Mamba achieved higher performance (AUROC 97.6%, 95% CI: 96.4%, 98.5%) for septic delirium compared to Mandarin (AUROC 92.7%, 95% CI: 90.3%, 95.0%). However, Mandarin achieved higher AUROCs in all other delirium phenotypes compared to baselines.

For delirium and coma subtypes, Mandarin and Mamba achieved the highest AUROCs in both external and prospective cohorts for the 12-hour horizon. Notably, both models maintained robust performance in the external cohort for hyperactive delirium (Mandarin: 72.5%, 95% CI: 65.6%, 77.7%; Mamba: 73.6%, 95% CI: 69.3%, 78.9%) compared to Neurological Scores (54.6%, 95% CI: 46.8%, 62.6%), and hypoactive delirium (Mandarin: 75.6%, 95% CI: 74.3%, 76.9%; Mamba: 75.3%, 95% CI: 74.1%, 76.8%) compared to CatBoost (63.5%, 95% CI: 62.0%, 65.4%). Given the absence of patients with hyperactive delirium in the prospective cohort, the performance for hypoactive delirium was the same as for overall delirium (as reported in Section 3.3). For coma subtypes, both Mandarin and Mamba achieved higher AUROCs for miscellaneous coma in external (Mandarin: 85.1%, 95% CI: 82.4%, 86.7%; Mamba: 84.3%, 95% CI: 82.3%, 86.0%) and prospective (Mandarin: 91.0%, 95% CI: 83.2%, 96.6%; Mamba: 92.2%, 95% CI: 87.7%, 96.7%) cohorts compared to both Neurological Scores (External: 68.9%, 95% CI: 65.8%, 71.3%; Prospective: 65.2%, 95% CI: 52.3%, 76.7%) and CatBoost (External: 68.9%, 95% CI: 66.5%, 72.3%; Prospective: 78.8%, 95% CI: 68.4%, 86.8%). A comprehensive comparison of all models at all time-horizons can be found in the Supplemental Material.

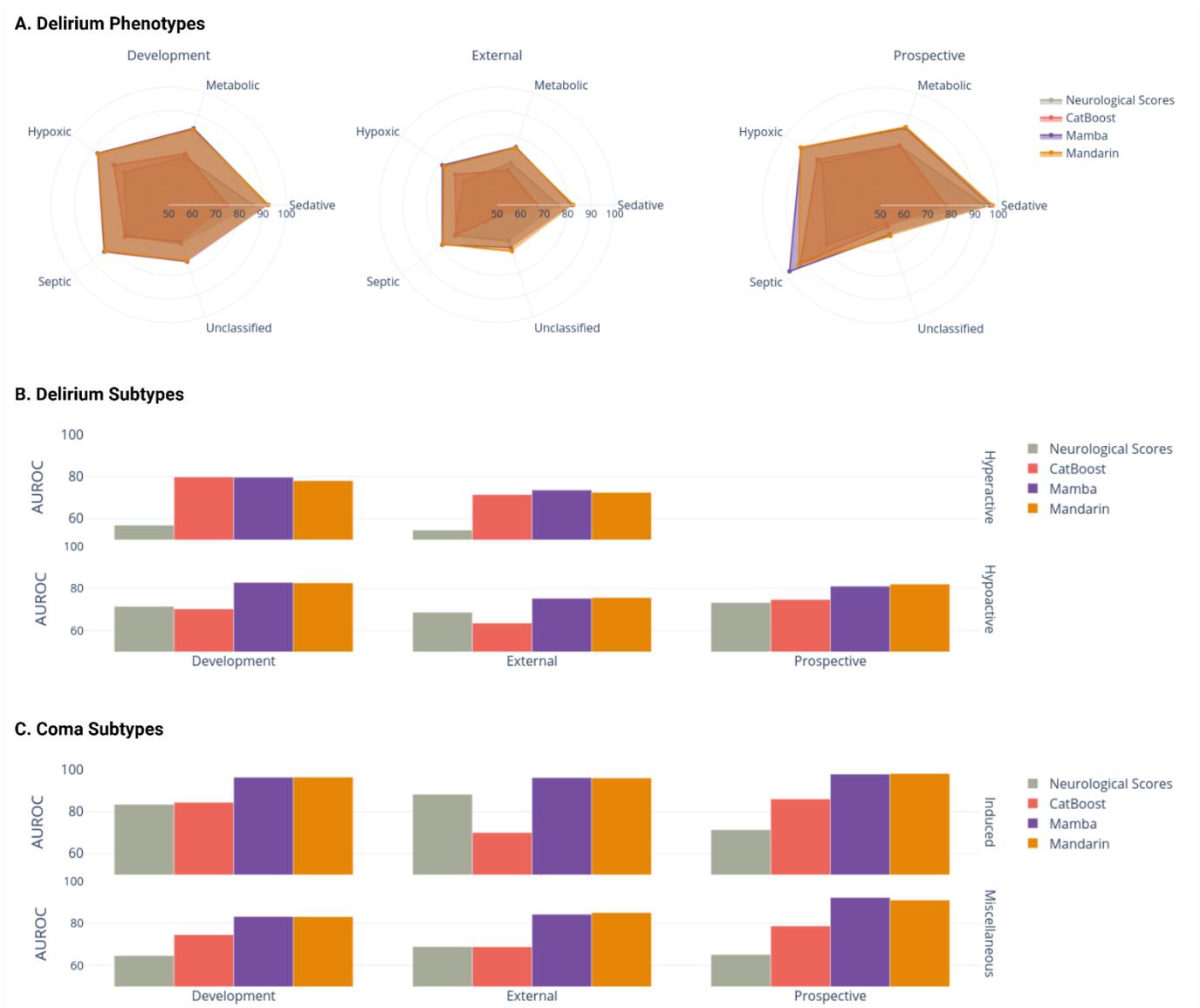

**Fig. 6: Model performance for predicting delirium and coma sub-phenotypes across different cohorts 12 hours in advance when current status is No ABD.** (A) Radar plots showing the predictive performance in terms of AUROC of Mandarin and baselines across validation cohorts for five delirium sub-phenotypes: Metabolic, Hypoxic, Septic, Sedative, and Unclassified. (B) Bar plots showing predictive performance in terms of AUROC for each model across different cohorts for two delirium subtypes: Hyperactive and Hypoactive. (C) Bar plots showing predictive performance in terms of AUROC for each model across different cohorts for two coma subtypes: Induced and Miscellaneous. Note: the prospective cohort had no patients with hyperactive delirium and, therefore, performance could not be computed.

3.5 Feature importance

The feature importance analysis identified key predictors of delirium, coma, and deceased in both external and prospective cohorts (Fig. 7). Neurological assessments, including CAM, GCS components, and

RASS, were strong predictors of delirium in both cohorts. Metabolic and inflammatory markers such as anion gap, blood urea nitrogen (BUN), white blood cell (WBC) count, and C-reactive protein (CRP) also played significant roles. In the prospective cohort, functional scores—Braden nutrition, activity, and mobility—along with lactic acid and arterial pH, emerged as additional key factors.

For coma, GCS components remained among the top predictors, along with anion gap, BUN, aspartate aminotransferase (AST), and international normalized ratio (INR). Fluid balance and metabolic markers, including serum sodium, platelet count, and urine specific gravity, were also influential. The prospective cohort further highlighted Braden activity and mobility scores, ionized calcium, and total protein. Mortality was strongly associated with neurological, metabolic, and hemodynamic factors. BUN, GCS components, RASS, anion gap, INR, and arterial pH ranked highly across both cohorts, while the prospective cohort further emphasized functional assessments (Braden scores), lactic acid, and CRP.

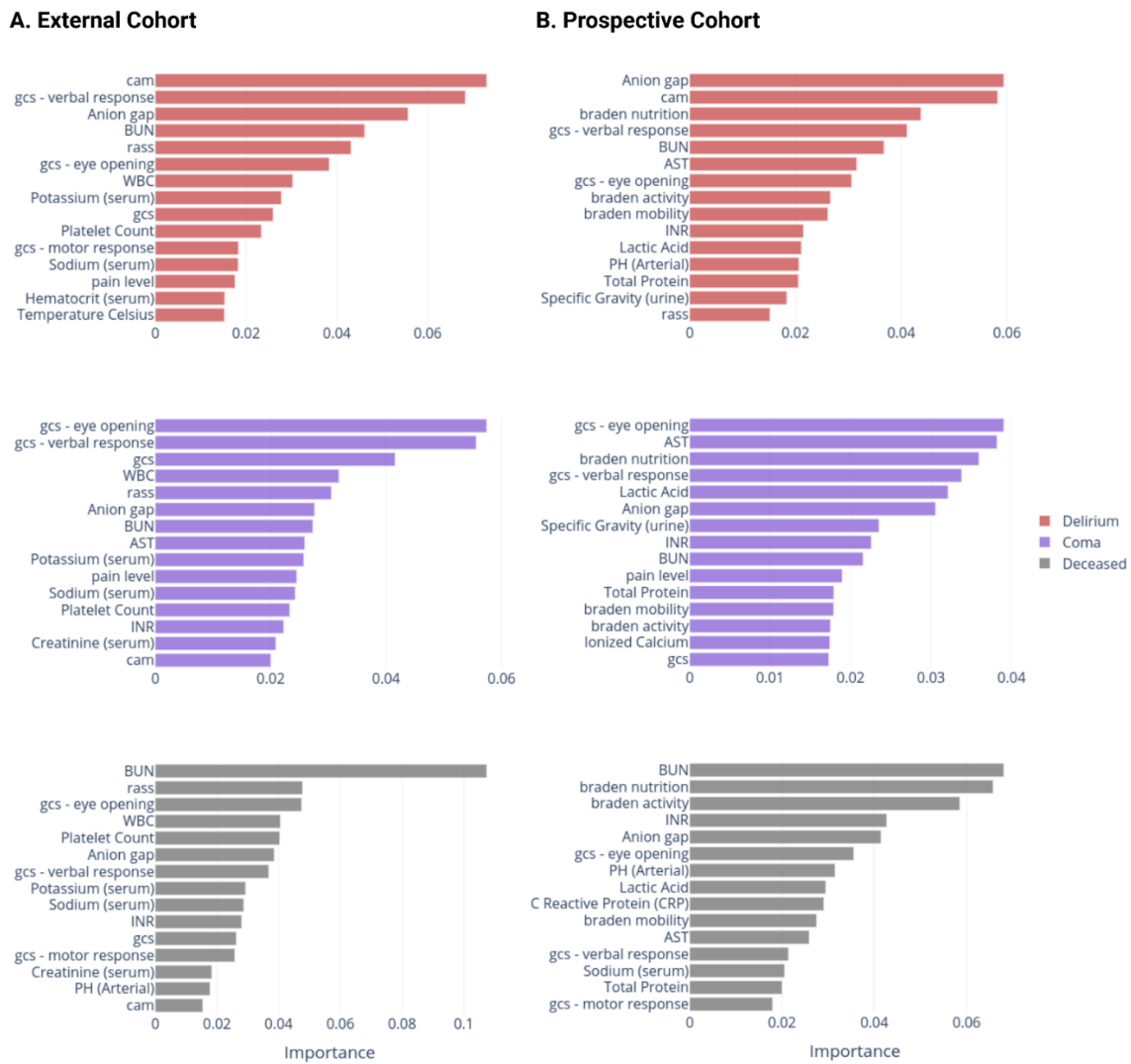

**Fig. 7: Feature importance of MANDARIN for predicting delirium, coma, and deceased across different cohorts and prediction windows when current status is No ABD.** Bar plots depicting feature importance as computed using integrated gradients for delirium (red), coma (purple), and deceased (grey) in the external (A) and prospective (B) validation cohorts.

# 4. Discussion

4.1 Main Findings and Interpretation

The results presented in this study show the development and validation of MANDARIN as a tool for real-time prediction of ABD in the ICU. To the best of our knowledge, this is the first study to continuously predict ABD (delirium, coma, and mortality) in the ICU by leveraging temporal and static EHR data. Previous studies have mostly focused on predicting delirium only, with varying performance. A systematic review of ICU delirium prediction models revealed that AUROC varied greatly (62.0%-94.0%) across 23 prediction models [20]. However, most of these studies have focused on single-point delirium prediction (*i.e.,* predicting whether the patient would develop delirium or not during their ICU admission). Few studies have developed dynamic delirium prediction models [22], [23], but have several limitations such as lack of external and prospective validation and not considering the brain status of the patient at the time of prediction. Similarly, studies developing ABD prediction models have also shown limitations such as small study cohorts and lack of external validation [24], [25]. While drawing direct comparisons between the performance of MANDARIN and these models is challenging due to the variation in development and validation settings, the results reported in this paper are comparable and even outperform what is seen in the literature. Notably, this study trains and validates the model on over 100,000 patients, the largest cohort among these studies, achieving strong performance in an external validation set spanning 15 hospitals and in a prospective validation. Furthermore, MANDARIN leverages state-of-the-art (SOTA) AI techniques with the use of a mixture-of-experts block for time series classification, enabling prediction of ABD across different time horizons and handling changes in patient brain status through a multi-branch approach.

When assessing performance for prediction of different delirium and coma phenotypes and subtypes, MANDARIN demonstrated robust performance compared to baselines across both external and prospective cohorts, demonstrating its ability to capture diverse manifestations of ABD more effectively. Traditional screening methods, such as GCS, CAM, and RASS, often struggle with the heterogeneity of ABD, particularly in atypical or overlapping presentations of delirium and coma [43]. For instance, the GCS has been noted to have practical limitations in daily routine assessments, especially in patients with non-neurological primary diseases, as it may not account for subtle derangements of consciousness like delirium [44]. Similarly, while the CAM-ICU is a validated tool for delirium screening, it may exhibit reduced sensitivity for identifying hypoactive delirium, underscoring the need for comprehensive clinical evaluations [45]. The superior performance of MANDARIN across different phenotypes suggests that deep learning models leveraging multiple data types (*i.e.,* vital signs, laboratory tests, medications) can better account for the complex pathophysiology underlying ABD compared to manual assessments.

The feature importance analysis showed the multifaceted nature of predicting delirium, coma, and mortality in ICU patients, highlighting the significance of integrating neurological assessments, metabolic and inflammatory markers, and functional scores. In particular, features such as anion gap, BUN, arterial pH, and potassium can indicate presence of metabolic acidosis, while WBC and platelet count can indicate inflammation due to infection, both of which have been associated to delirium [46]. Furthermore, anemia (hematocrit and Braden nutrition score), immobilization (Braden activity and mobility scores), and certain biomarkers (CRP) have all been associated as precipitating factors of delirium [47]. In predicting coma and mortality, key features also included GCS and metabolic markers, which can reflect derangements that can adversely affect neurological status and are associated with increased mortality [48].

The prospective evaluation of MANDARIN on ICU patient data acquired with a real-time platform provides a validation method of the potential application of this model in the real world. The performance seen in this cohort was comparable to the other validation cohorts, which shows MANDARIN can be readily deployed and employed in real-time to assist clinicians in decision-making.

4.2 Limitations and Future Work

Some of the limitations of this study include the challenge of labeling brain status continuously in the ICU. Neurological assessments are performed intermittently in the ICU [7], which can lead to missing many altered mental status changes [8]. The lack of consistent and accurate assessments limits the model's robustness in a real-world setting. The small number of features can limit model performance as well. Large models can usually benefit from a large number of features. The present study used a limited subset of features common in all three databases (UFH, MIMIC, and eICU). Future studies will leverage use of Large Language Models (LLMs) to represent EHR data in text form [49], removing the need of mapping features. Furthermore, although a prospective validation was conducted, such validation was conducted retrospectively. Deployment of the model in real-time and a clinical adjudication are necessary to validate the model's real-world performance fully.

## 5. Conclusion

This study presents MANDARIN, the first real-time prediction model for ABD in the ICU, integrating temporal and static EHR data. Unlike previous models, MANDARIN dynamically predicts delirium, coma, and mortality with superior performance across external and prospective validation cohorts. By leveraging a mixture-of-experts framework, MANDARIN outperforms traditional screening tools (GCS, CAM, RASS) in capturing ABD's complexity. Prospective evaluation supports its clinical applicability, though challenges remain, including intermittent neurological assessments and feature limitations. Future work will explore the use of LLMs for enhanced EHR representation and real-time clinical deployment to improve early detection and patient outcomes.

## 6. Acknowledgement

A.B, P.R., and T.O.B. were supported by NIH/NINDS R01 NS120924, NIH/NIBIB R01 EB029699. PR was also supported by NSF CAREER 1750192.

## 7. Data & Code Availability

The data from MIMIC and eICU is publicly available and can be accessed through PhysioNet at the following links after obtaining approval: MIMIC-IV (https://physionet.org/content/mimiciv/2.2/) and eICU-CRD (https://physionet.org/content/eicu-crd/2.0/). The data from UFH is private and has not been approved for public use.

All analyses carried out in this study were performed using Python version 3.10.8. The Python package torch version 2.1.2, mamba-ssm 1.1.1, and scikit-learn 1.4.0 was used for developing the machine learning and deep learning models. Integrated gradients were implemented using the Python package Captum version 0.7.0. Visualization plots/graphs were obtained using Python packages matplotlib version 3.8.3 and seaborn version 0.13.2. Statistical analyses were performed using scipy version 1.12.0.